\documentclass[runningheads]{llncs}
\usepackage{eccv}

\makeatletter
\newcommand\myparagraph{\@startsection{paragraph}{4}{\z@}%
  {0.5ex plus 0.2ex minus .1ex}%
  {-1em}%
  {\normalfont\normalsize\itshape}}
\makeatother
\usepackage{eccvabbrv}
\usepackage{graphicx}
\usepackage{booktabs}
\usepackage[accsupp]{axessibility}
\usepackage{hyperref}
\usepackage{orcidlink}
\usepackage{amsmath,amsfonts,amssymb}
\usepackage{bm}
\usepackage{textcomp}
\usepackage{algorithm}
\usepackage{algorithmic}
\usepackage{tabularx}     
\usepackage{booktabs}     
\usepackage{multirow}    
\usepackage{adjustbox}
\usepackage{makecell}     
\usepackage{siunitx}      
\usepackage{array}        
\usepackage{colortbl}     
\usepackage{graphicx}      
\usepackage{stfloats}      
\usepackage{placeins}      
\usepackage[dvipsnames,table]{xcolor}
\usepackage{tikz}

\usepackage{tcolorbox}
\usepackage{listings}
\usepackage{inconsolata}
\usepackage{xcolor}
\usepackage{url}
\usepackage{verbatim}

\usepackage{caption}

\newcolumntype{C}{>{\centering\arraybackslash}X}
\usetikzlibrary{arrows.meta, positioning, fit}
\definecolor{textcolor}{RGB}{255,140,0}
\definecolor{imagecolor}{RGB}{70,130,180}
\definecolor{crosscolor}{RGB}{138,43,226}
\definecolor{losscolor}{RGB}{220,20,60}

\tikzset{
  module/.style={rectangle, rounded corners, draw=black, thick, align=center, minimum width=3cm, minimum height=1cm},
  textmod/.style={module, fill=textcolor!15},
  imgmod/.style={module, fill=imagecolor!15},
  crossmod/.style={module, fill=crosscolor!15},
  lossmod/.style={rectangle, draw=losscolor, thick, dashed, minimum width=2.2cm, minimum height=0.8cm, align=center, font=\scriptsize},
  line/.style={-Latex, thick},
  dashedbox/.style={draw=black, thick, dashed, rounded corners, inner sep=0.3cm}
}

\begin{document}

\title{Learning to Compose: Revisiting Proxy Task Design for Zero-Shot Composed Image Retrieval}
\titlerunning{FoCo: Focus-then-Complete for ZS-CIR}

\author{Jingjing Zhang\orcidlink{0000-0003-4819-2585} \and
Lei Zhang\thanks{Corresponding author.} \and
Zheren Fu \and
Zhendong Mao}

\authorrunning{J.~Zhang et al.}

\institute{University of Science and Technology of China, Hefei, China\\
\email{zjj1029@mail.ustc.edu.cn, \{leizh23, fzr, zdmao\}@ustc.edu.cn}}

\maketitle

\begin{abstract}
Composed Image Retrieval (CIR) retrieves a target image from a reference image and a textual modification. 
While supervised CIR relies on costly triplets, Zero-Shot CIR (ZS-CIR) alleviates this reliance through proxy tasks trained on image–text pairs.
However, existing proxy tasks primarily enhance visual and textual representations to accommodate a predefined composition mechanism such as pseudo-word injection into a frozen text encoder or linear feature arithmetic. 
As a result, the composition function itself remains unlearned, limiting the model’s ability to express diverse and fine-grained semantic modifications.
To address this, we propose FoCo, which models composition as two coordinated stages: focusing on modification-relevant visual content, and then completing the target semantics.
We realize these through two proxy tasks: \textit{text-anchored visual aggregation} to selectively gather visual content guided by localized textual semantics, and \textit{context-conditioned semantic completion} to transform these aggregated visuals with the remaining scene context into a coherent composed representation.
The tasks are trained jointly with a cross-instance contrastive objective, encouraging semantic diversity and discouraging shortcut composition strategies. 
Extensive experiments on four ZS-CIR benchmarks show FoCo’s state-of-the-art performance and improved generalization.
\end{abstract}

\section{Introduction}
\label{sec:intro}

\begin{figure}[t]
    \centering
    \includegraphics[height=4.5cm]{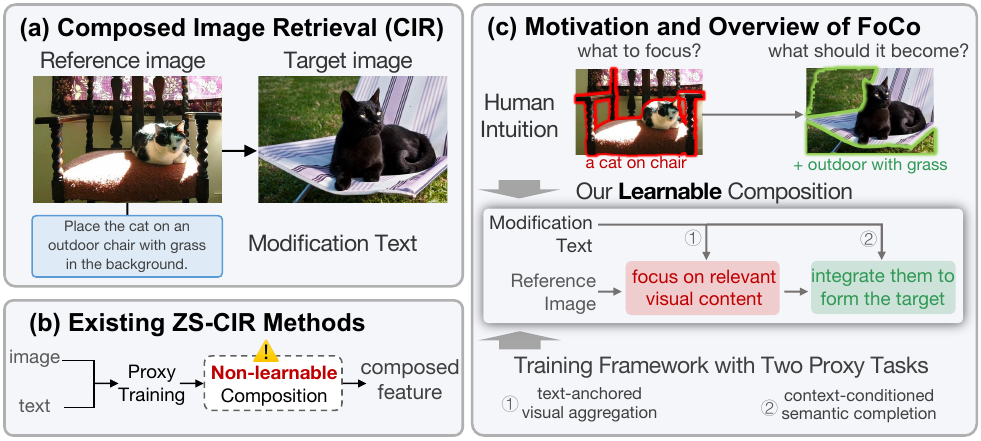}
    \caption{
    (a) The CIR task retrieves a target image from a reference image and a modification text.
    (b) Existing Zero-Shot CIR uses proxy tasks to avoid supervised triplets, but relies on fixed, non-learnable composition rules, causing the proxy learner to mainly adapt features to these rules.
    (c) Our learnable composition framework is inspired by human-like selective composition: first focus on the visual semantics relevant to the modification, 
    and then integrate them with the text to form the target representation.
    }
    \label{fig:intro}
\end{figure}

Composed Image Retrieval (CIR)~\cite{delmas2022artemis, wen2023target} aims to retrieve a target image using a reference image and a textual modification, enabling users to specify how the reference should be changed rather than describing the target from scratch.
Recently, Zero-Shot CIR (ZS-CIR)~\cite{saito2023pic2word} has emerged as an important paradigm that eliminates the need for triplet annotations (reference image, textual modification, target image) during training, making the task both practical and challenging.

Most existing ZS-CIR approaches design proxy tasks on pretrained vision--language models to avoid triplet supervision. 
However, these methods mainly focus on learning feature representations that conform to a predefined, non-learnable composition mechanism, leaving the composition process itself unmodeled. 
Two prevalent paradigms illustrate this limitation.
Some methods project image features into pseudo-words and compose them with a frozen text encoder~\cite{ tang2024context, bai2024sentence, gu2024language, li2025rethinking, li2025hierarchy}.
Such encoders are pretrained to maximize global image–text semantic alignment, rendering them ill-suited to capturing the localized semantic changes central to CIR.
Others rely on simple vector arithmetic for composition~\cite{jang2024spherical, chen2025pretrain}, which is fundamentally limited in representing the non-linear semantic transformations required in real-world modifications.
In essence, these methods tailor representations to a fixed composition rule, whereas CIR calls for a learnable composition mechanism that, by being learned from data, can naturally adapt to diverse, context-dependent modifications. 
Motivated by human cognitive processes~\cite{oliva2007role}, we observe that humans interpret scenes by first directing attention to regions of interest and then using broader contextual cues to refine their understanding.
This principle motivates modeling CIR composition as a two-stage procedure, as illustrated in~\cref{fig:intro}(c): first identifying modification-relevant semantics in the reference image, 
then integrating them with the textual instruction to form the target representation.
Accordingly, we propose \textbf{Focus-then-Complete (FoCo)}, a ZS-CIR framework with a learnable composition mechanism.
CIR textual modifications typically refer to specific objects, attributes, or background elements, yet such fine-grained semantics are not explicitly reflected in standard image–caption pairs.
To expose the model to these diverse semantic aspects, we decompose each caption into local captions describing distinct components and a contextual caption capturing the remaining scene information.
This caption decomposition naturally supports the focus–complete process.
Leveraging these data, we design two proxy objectives that induce the two capabilities required by FoCo.
The first, \textbf{Text-Anchored Visual Aggregation}, selectively gathers visual evidence conditioned on each local caption to obtain a localized visual–semantic representation.
The second, \textbf{Context-Conditioned Semantic Completion}, 
transforms each localized representation with the corresponding contextual caption to recover the full semantics.
Together, these objectives allow FoCo to learn where to focus and how to complete.
Training these modules, however, is nontrivial.
Caption decomposition yields multiple semantic views of an image, each paired with a contextual caption.
While these multiple semantic cues are essential for learning the composition mechanism, they also make the optimization considerably more challenging than standard image–text contrastive learning.
A standard contrastive objective assumes a single positive caption per image and constructs negatives through coarse instance-level mismatches, which are too weak to supervise such fine-grained distinctions and often lead to unimodal shortcut behavior.
To address this, we design a \textbf{Cross-Instance Contrastive Training} scheme that leverages multiple positives and carefully constructed negatives to properly supervise learning from these multiple semantic cues introduced by caption decomposition.
This formulation encourages accurate alignment between each local caption and its visual evidence while reducing shortcut behavior.

Our contributions are fourfold:
(1) We propose Focus-then-Complete (FoCo), a ZS-CIR framework that models composition as a learnable process aligned with human cognition.
(2) We introduce two proxy objectives with a caption decomposition strategy, enabling semantic focusing and context-aware completion without triplet supervision.
(3) We develop a cross-instance contrastive scheme that uses multiple positives and cross-instance negatives to improve localized grounding and avoid shortcut learning.
(4) FoCo delivers consistent gains across ZS-CIR benchmarks, demonstrating strong effectiveness and robustness.

\section{Related Work}
The core challenge of composed image retrieval (CIR) lies in identifying how the modification text should alter the reference image, and encoding this change while preserving the unchanged visual content~\cite{sun2025leveraging, li2025encoder}. Early methods mainly adopt generic fusion mechanisms. TIRG~\cite{vo2019composing} introduces a gated residual connection to modulate visual features using textual input, and MAAF~\cite{dodds2020modality} employs residual attention fusion to guide retrieval through cross-modal interactions. To achieve more controllable and semantically guided composition, later approaches design dedicated modules for integrating the reference and the textual modification. CoSMo~\cite{lee2021cosmo} decomposes textual features into content and style components to support targeted modulation. TMCIR~\cite{wang2025tmcir} introduces token-level modality balancing to better capture fine-grained user intent. DetailFusion~\cite{yang2025detailfusion} performs hierarchical adaptive fusion to preserve both coarse and detailed cues. PAIR~\cite{fu2025pair} separates coherent and incoherent relations between the image and the text to enable more controlled semantic integration. 
This line of work~\cite{yang2024decomposing, tian2025ccin} consistently highlights the importance of explicitly modeling how multimodal signals interact during composition. However, all these approaches rely on triplet-level supervision, which limits scalability and constrains semantic diversity due to the small size of existing CIR datasets. These limitations have motivated the development of ZS-CIR, which removes this dependency and enables training with large-scale image–caption pairs. A key challenge in this setting is designing proxy tasks that mitigate the distribution mismatch between training inputs (image–caption pairs) and inference inputs (image–text modifications). In this context, we focus on building a learnable composition mechanism that can reliably combine the visual and textual inputs under the ZS-CIR paradigm.

\section{Method}
We address Zero-Shot CIR using only paired image–text data. 
To this end, we propose Focus-then-Complete (FoCo), a self-supervised framework that treats visual–textual composition as a two-stage procedure: it first extracts the visual semantics relevant to the textual instruction, 
and then transforms the focused representation under textual guidance to yield the target semantics.
This two-stage formulation is supported by decomposed caption signals introduced in~\cref{sec:local-sampling}, which provide supervision tailored to each stage. 
Concretely, FoCo consists of \textit{text-anchored visual aggregation} to gather modification-relevant semantics and \textit{context-conditioned semantic completion} 
to transform them into a target representation under textual guidance.

\subsection{Local Caption Generation}
\label{sec:local-sampling}
\begin{figure*}[t]
\centering
\includegraphics[width=0.95\textwidth]{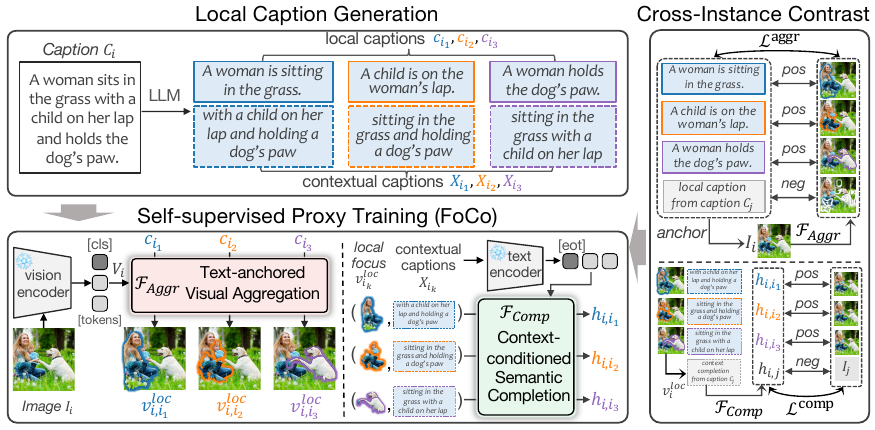}
\caption{
Overview of the proposed Focus-then-Complete (FoCo) framework.
\textbf{Local Caption Generation} decomposes each global caption $C_i$ into local captions $c_{i_k}$ with contexts $X_{i_k}$.
In the \textbf{Self-supervised Proxy Training} stage,
Text-anchored Visual Aggregation ($F_{\text{Aggr}}$) extracts localized visual features $v^{\text{loc}}_{i,k}$ guided by each $c_{i_k}$, and
Context-conditioned Semantic Completion ($F_{\text{Comp}}$) transforms $v^{\text{loc}}_{i,k}$ with $X_{i_k}$ to yield composed representations $h_{i,k}$.
A \textbf{Cross-instance Contrastive Objective} treats all representations derived from the same image $I_i$ as positives and uses those from other images $I_j$ as negatives to prevent shortcut learning.
FoCo thus learns an effective two-stage composition mechanism from image--caption data without triplet supervision.
}
\label{fig:framework}
\end{figure*}

Most vision–language models (VLMs) align images and captions primarily at a global level, capturing overall correspondences between text and the visual scene but lacking the ability to model fine-grained linguistic elements and localized visual cues~\cite{fineclip}.
However, textual modifications in CIR often correspond to \textit{localized semantic changes}, such as adding an object, changing an action, or modifying the scene. Global captions, which entangle multiple semantic components, are therefore insufficient for supervising fine-grained composition.
To address this, we introduce \textit{Local Caption Generation}, which uses a large language model (LLM), \textit{Llama-3.1-8B-Instruct}~\cite{grattafiori2024llama}, to semantically decompose each global caption $C_i$ into a set of coherent \emph{local captions} $\{c_{i_k}\}_{k=1}^{K}$, where $K$ denotes the number of local captions for image $i$. 
Each local caption describes a distinct semantic component of the scene, capturing fine-grained textual cues that align with localized visual content. 
As illustrated in \cref{fig:framework}, these local captions provide the supervision signals for our proxy tasks and correspond to the focus stage of FoCo.
Alongside each $c_{i_k}$, 
the LLM also generates a \textit{contextual caption} $X_{i_k}$, which contains the scene information beyond $c_{i_k}$ from the original caption.
Together, each pair $\langle c_{i_k}, X_{i_k} \rangle$ covers the full semantics of the scene from different perspectives:
$c_{i_k}$ serves as the focus anchor for visual aggregation, while $X_{i_k}$ provides the transformation condition for semantic completion. 
The prompt used for generation is provided in the Appendix.

\subsection{Proxy Tasks for Aggregation and Completion}
\label{sec:proxy_tasks}
To equip the model with localized semantic focusing and contextual completion, we design two proxy tasks built upon the generated local captions and their corresponding contextual captions described in \cref{sec:local-sampling}. 
Each local caption acts as a \textit{textual anchor}, providing the basis for visual aggregation, while its contextual caption guides subsequent semantic completion, transforming the aggregated visual cue into the target representation.
Together, these two tasks implement the overall ``focus-then-complete'' pipeline.

\myparagraph{\textbf{Text-Anchored Visual Aggregation.}}
\label{sec:text-anchored-aggregation}
CIR requires reasoning about how specific semantic concepts vary across images, yet global image representations often obscure subtle differences tied to object- or region-level details. 
We therefore treat each local caption as a textual anchor that conditions the model to dynamically aggregate the relevant visual content. Formally, given an image $I_i$ with local captions $\{c_{i_k}\}_{k=1}^K$, this module generates localized visual features conditioned on each anchor:
\begin{equation}
v_{i,i_k}^{\text{loc}} = F_{\text{Aggr}}(c_{i_k}, I_i),
\end{equation}
where $F_{\text{Aggr}}(\cdot)$ is the text-anchored aggregation operator. 
The image $I_i$ is represented by a global feature $v_i^g$ and a set of patch features $\mathcal{V}_i^p$ extracted from the vision backbone. 
Each anchor $c_{i_k}$ is fused with $v_i^g$ through a lightweight MLP to form a text-guided query:
\begin{equation}
q_{i_k} = \mathrm{MLP}(c_{i_k} \odot v_i^g),
\end{equation}
where $\odot$ denotes element-wise multiplication. 
Here, $q_{i_k}$ is conditioned on both textual and global visual context, enabling anchor-specific attention over the visual feature space. 
The patch features $\mathcal{V}_i^p$ serve as keys and values in a cross-attention operation:
\begin{equation}
\label{eq:focus_attn}
v_{i,i_k}^{\text{loc}} = \mathrm{CrossAttn}(q_{i_k}, \mathcal{V}_i^p).
\end{equation}
In practice, it is implemented with a multi-head cross-attention layer.
We use $i$ to index images, $k$ to index local captions of image $i$, and $j \neq i$ to index other images. For clarity, we denote $v^{\text{loc}}_{i,j_{k'}}$ as the localized feature of image $I_i$ conditioned on the local caption $c_{j_{k'}}$ derived from $C_j$.

\myparagraph{\textbf{Context-Conditioned Semantic Completion.}}
Building on the aggregation process, this module transforms the localized visual evidence $v_{i,i_k}^{\text{loc}}$ under the guidance of a contextual caption $X_{i_k}$ to complete the target semantics.

Formally, we construct a composite input sequence that aligns with the composition logic: the textual context $X_{i_k}$ serving as a transformation instruction, the localized visual feature $v_{i,i_k}^{\text{loc}}$ representing the subject to be transformed, and the remaining visual features $\bar{v}_{i,i_k}$ as the background.
We apply modality-specific mapping networks to project these components and concatenate them along the sequence dimension:
\begin{equation}
S_{i,k} = [\,f_t(X_{i_k})\,;\, f_v(v_{i,i_k}^{\text{loc}})\,;\, f_v(\bar{v}_{i,i_k})\,],
\end{equation}
where $\bar{v}_{i,i_k}$ denotes the background representation, obtained by aggregating visual patch features weighted by the inverse of the attention map derived in \cref{eq:focus_attn}.
This design ensures the dynamic incorporation of unattended regions as visual background during the composition process.

We model the composition as a text-guided transformation process.
Specifically, we initialize a learnable query token with the localized visual feature $v_{i,i_k}^{\text{loc}}$, establishing it as the semantic anchor for transformation.
This query token interacts with the sequence $S_{i,k}$ through multi-layer Transformer blocks $\mathcal{T}(\cdot)$, dynamically attending to the transformation instructions in $X_{i_k}$ and the background details in $\bar{v}_{i,i_k}$ to generate the target representation:
\begin{equation}
h_{i,i_k} = \mathcal{T}(q=v_{i,i_k}^{\text{loc}}, kv=S_{i,k}).
\end{equation}
The final output $h_{i,i_k}$ encapsulates a text-guided transformation of the localized visual anchor. 
The contextual caption thus acts as a training proxy for the modification text encountered at inference, guiding the transformation of the focused visual subject toward the intended target semantics.

Although the two modules form a coherent composition process, training them effectively remains challenging, 
as it requires robust cross-modal reasoning across diverse contexts. 
We therefore adopt a unified \textit{cross-instance semantic contrastive} framework.

\subsection{Cross-Instance Semantic Contrastive Training}
\label{sec:contrastive_training}
To enable fine-grained semantic grounding and context-aware completion for ZS-CIR, we move beyond standard contrastive learning.
We propose a cross-instance semantic contrastive framework with two synergistic objectives:
an \textit{aggregation loss} ($\mathcal{L}^{\text{aggr}}$) for grounding local concepts and a \textit{completion loss} ($\mathcal{L}^{\text{comp}}$) 
for supervising the transformation toward globally consistent target semantics.
This framework leverages the decomposed local captions and their contextual captions from~\cref{sec:local-sampling}
to construct challenging positive and negative pairs that enforce robust cross-modal representations.

\myparagraph{\textbf{Aggregation Objective.}}
Given an image $I_i$, the aggregation operator $F_{\text{Aggr}}$ produces, for each local caption $c_{i_k}$, a \textit{text-conditioned visual representation} $v^{\text{loc}}_{i,i_k}$ that selectively captures the visual content grounded to that local description.
Since each anchor highlights a distinct semantic aspect of the image, the resulting pairs $\langle v^{\text{loc}}_{i,i_k},\, c_{i_k} \rangle$ naturally form multiple positive samples within the same image. To effectively learn from these multiple positives, we adopt the sigmoid-based contrastive loss from~\cite{zhai2023sigmoid}, which accommodates multi-positive supervision and provides more stable optimization than softmax-based InfoNCE.

A critical aspect of our framework lies in the construction of \textit{negative pairs}, which play a key role in preventing \textit{shortcut learning}. 
To illustrate this concretely, consider a naive negative sampling strategy where 
a visual feature is aggregated from image $I_i$ using an anchor $c_{j_{k^{\prime}}}$ from another sample. 
This feature is then contrasted with the anchor $c_{i_k}$, \ie, $\langle v_{i,j_{k^{\prime}}}^{\text{loc}}, c_{i_k}\rangle$. 
In this setup, the aggregation module may degenerate by merely encoding the semantics of the conditioning text $c_{j_{k^{\prime}}}$, ignoring the visual evidence from $I_i$. 
As a result, the model simply distinguishes pairs by comparing the textual similarity between $c_{j_{k^{\prime}}}$ and $c_{i_k}$ 
without genuinely producing the desired text-conditioned visual feature $v_{i,j_{k^{\prime}}}^{\text{loc}}$. 

To avoid this degeneration, we formulate our negative pairs as $\langle v_{i,j_{k^{\prime}}}^{\text{loc}}, \! c_{j_{k^{\prime}}} \rangle$. 
These negatives force the model to discern whether $v_{i,j_{k^{\prime}}}^{\text{loc}}$ is truly supported by the visual evidence in image $I_i$, rather than simply relying on the anchor text $c_{j_{k^{\prime}}}$. 
This ensures meaningful contributions  to the contrastive learning objective from both modalities.
The aggregation loss for sample $i$ is defined as:
\begin{align}
\!\! \mathcal{L}^{\text{aggr}}_i =
-\Big[
\sum_{k=1}^{K} 
\phi^{+}\!\big(v_{i,i_k}^{\text{loc}}, c_{i_k}\big)
+\sum_{j\neq i} 
\phi^{-}\!\big(v_{i,j_{k^{\prime}}}^{\text{loc}}, c_{j_{k^{\prime}}}\big)
\Big],
\label{eq:agg_loss}
\end{align}
where $k^\prime$ indexes a randomly sampled local caption from the global caption~$C_j$, and $\phi^{\pm}(\cdot,\cdot)$ denotes the sigmoid-based scoring function~\cite{zhai2023sigmoid}, defined as
\begin{equation}
\phi^{\pm}(u,v) = \log \sigma\!\left(\pm\,\tfrac{\operatorname{sim}(u,v)}{\tau}\right),
\end{equation}
where $\operatorname{sim}(u,v)$ is the cosine similarity, and $\tau$ is a temperature scaling factor.

\myparagraph{\textbf{Completion Objective.}}
As with the aggregation objective in \cref{eq:agg_loss}, we extend the cross-instance contrastive principle to the context-guided transformation process.
Specifically, the composition module $F_{\text{Comp}}$ yields a composed representation $h_{i,i_k}$ by transforming the localized visual feature $v^{\text{loc}}_{i,i_k}$ conditioned on its contextual caption $X_{i_k}$.
Each $h_{i,i_k}$ is aligned with the corresponding global visual embedding $v_i^g$, forming multi-perspective positive pairs that capture context-conditioned semantic completion.
For negatives, we construct cross-instance pairs by replacing the text with one from another sample, yielding the contrastive pair $\langle h_{i,j_{k^{\prime}}}, v_j^g \rangle$. 
This construction intentionally disrupts semantic coherence by combining a localized visual cue from $I_i$ with an irrelevant linguistic context. 
This ensures that both visual and textual modalities contribute effectively to the completion objective, preventing shortcut learning in a manner similar to \cref{eq:agg_loss}. 
The loss is formulated as:
\begin{align}
\!\! \mathcal{L}^{\text{comp}}_i \!=
-\big[
\sum_{k=1}^{K} 
\phi^{+}\!\left(h_{i,i_k}, v_i^g\right)
+\sum_{j\neq i} 
\phi^{-}\!\left(h_{i,j_{k^{\prime}}}, v_j^g\right)
\big].
\label{eq:comp_loss}
\end{align}

\myparagraph{\textbf{Discussion.}}
Our contrastive design brings two key benefits. 
(i) \emph{Multi-fragment positives} provide simultaneous supervision over multiple localized correspondences within a single image–text pair, strengthening localized grounding without intra-pair competition.
(ii) \emph{Cross-instance negatives} couple text from one sample with visual evidence from another, compelling joint reasoning across modalities and suppressing unimodal shortcuts.
Together, these mechanisms align the aggregation and completion modules: the model learns to activate the right visual evidence under language control and to validate global semantic coherence under visual grounding, producing representations that are both locally precise and globally consistent for ZS-CIR.

\myparagraph{\textbf{Overall Training Objective.}}
The two objectives act in synergy: $\mathcal{L}^{\text{aggr}}$ learns which visual evidence is anchored by each local caption, 
while $\mathcal{L}^{\text{comp}}$ enforces consistency between the transformed representation and the full image semantics.
The final training objective is the sum of the aggregation and completion losses, computed over all samples in the mini-batch.
\begin{align}
\mathcal{L} = \sum_i \left( \mathcal{L}^{\text{aggr}}_i + \mathcal{L}^{\text{comp}}_i \right),
\label{eq:final_loss}
\end{align} 
By jointly optimizing grounding and completion, the model learns visual–language representations that are both locally precise and globally coherent.

\subsection{Inference with FoCo}
\label{sec:inference}
At inference, we follow the ``focus-then-complete'' principle. The model first grounds the modification text $T_m$ within the visual context of the reference image $I_r$, producing a localized feature $v_r^{\text{loc}} = F_{\text{Aggr}}(T_m, I_r)$ that highlights the regions relevant to the modification. 
This feature is then passed to the completion module, which transforms it under the modification instruction to produce the composed representation $h_r = F_{\text{Comp}}(v_r^{\text{loc}}, T_m)$.
Finally, $h_r$ is compared with the global embeddings ${v_j^g}$ of gallery images using cosine similarity to retrieve the target. This focus-then-complete pipeline enables FoCo to interpret modification instructions by first localizing their visual references and then composing the required semantic change.

Note that the default inference pipeline does not involve an LLM, and uses $T_m$ directly for both aggregation and completion. This design prioritizes inference efficiency and avoids external dependencies at test time. 
Optionally, an LLM can be further employed to decompose $T_m$ in line with the training protocol, thereby more fully exploiting the model's capacity for complex modifications. Detailed results of this variant are provided in Appendix Sec. C.
\newcolumntype{C}{>{\centering\arraybackslash}S[table-format=2.1]}
\newcommand{\conf}[1]{\textcolor{gray}{#1}}
\newcommand{\sepline}{{\color{darkgray!85}\vrule width 0.8pt}}
\sisetup{detect-weight=true, detect-inline-weight=math, input-symbols={-}}
\begin{table*}[t!]
  \centering
  \caption{Retrieval results on FashionIQ validation set. Best scores are in bold.}
  \renewcommand{\arraystretch}{0.9}
  \setlength{\tabcolsep}{2pt}
  \begin{adjustbox}{width=0.9\textwidth}
  \begin{tabular}{
    >{\centering\arraybackslash}p{1.8cm}
    l
    *{8}{C}
  }
    \toprule
    \multirow{4}{*}{\centering Backbones} & \multirow{4}{*}{\centering Methods} & \multicolumn{8}{c}{FashionIQ} \\
    \cmidrule(lr){3-10}
    & & \multicolumn{2}{c}{Dress} & \multicolumn{2}{c}{Toptee} & \multicolumn{2}{c}{Shirt} & \multicolumn{2}{c}{Average} \\
    \cmidrule(lr){3-4} \cmidrule(lr){5-6} \cmidrule(lr){7-8} \cmidrule(lr){9-10}
    & & \multicolumn{1}{c}{R10} & \multicolumn{1}{c}{R50} & \multicolumn{1}{c}{R10} & \multicolumn{1}{c}{R50} & \multicolumn{1}{c}{R10} & \multicolumn{1}{c}{R50} & \multicolumn{1}{c}{R10} & \multicolumn{1}{c}{R50} \\
    \midrule
    \multirow{9}{*}{\centering ViT-L/14}
    & Pic2Word~\cite{saito2023pic2word} \conf{CVPR'23} & 20.0 & 40.2 & 27.9 & 47.4 & 26.2 & 43.6 & 24.7 & 43.7 \\
    & SEARLE-XL~\cite{baldrati2023zero} \conf{ICCV'23} & 20.5 & 43.1 & 29.3 & 50.0 & 26.9 & 45.6 & 25.6 & 46.2 \\
    & Context-I2W~\cite{tang2024context} \conf{AAAI'24} & 23.1 & 45.3 & 30.6 & 52.9 & 29.7 & 48.6 & 27.8 & 48.9 \\
    & LinCIR~\cite{gu2024language} \conf{CVPR'24} & 20.9 & 42.4 & 28.8 & 50.2 & 29.1 & 46.8 & 26.3 & 46.5 \\
    & SlerpTAT~\cite{jang2024spherical} \conf{ECCV'24} & 23.4 & 45.1 & 32.0 & 51.2 & 29.9 & 46.5 & 28.4 & 47.6 \\
    & PrediCIR~\cite{tang2025missing} \conf{CVPR'25} & 25.4 & \textbf{49.5} & 33.1 & 55.4 & 31.8 & 52.0 & 30.1 & 52.3 \\
    & MoA~\cite{li2025rethinking} \conf{SIGIR'25} & 25.2 & 48.5 & 33.2 & 54.8 & 31.9 & 50.7 & 30.1 & 51.3 \\
    & HIT~\cite{li2025hierarchy} \conf{ICCV'25} & 25.6 & 47.1 & 32.8 & 54.7 & 32.4 & 51.2 & 30.3 & 51.0 \\
    & \textbf{FoCo} & \textbf{26.5} & 49.3 & \textbf{34.2} & \textbf{56.0} & \textbf{33.8} & \textbf{54.3} & \textbf{31.5} & \textbf{53.2} \\
    \midrule
    \multirow{3}{*}{\centering ViT-G/14}
    & LinCIR~\cite{gu2024language} \conf{CVPR'24} & 38.1 & 60.9 & 50.5 & 71.1 & 46.8 & 65.1 & 45.1 & 65.7 \\
    & PrediCIR~\cite{tang2025missing} \conf{CVPR'25} & 39.7 & 62.4 & 53.7 & \textbf{73.6} & 48.2 & 67.4 & 47.2 & 67.8 \\
    & \textbf{FoCo} & \textbf{41.1} & \textbf{64.6} & \textbf{54.6} & 72.2 & \textbf{51.5} & \textbf{69.2} & \textbf{49.1} & \textbf{68.7} \\
    \bottomrule
  \end{tabular}
  \end{adjustbox}
  \label{tab:fiq-only}
\end{table*}

\section{Experiments}

\subsection{Settings}
\noindent\textbf{Implementation Details.}
Following~\cite{tang2024context,gu2024language,tang2025missing}, FoCo uses frozen CLIP backbones, including ViT-L/14~\cite{radford2021learning} and ViT-G/14~\cite{ilharco_gabriel_2021_5143773}.
The CLIP image and text encoders remain fixed during training.  
Two lightweight trainable modules are introduced: the aggregation module $F_{\text{Aggr}}$ is implemented as a single-layer cross-attention block followed by a feed-forward projection, while the completion module $F_{\text{Comp}}$ comprises two Transformer blocks (each with 8-head attention and a feed-forward network) for integrating contextual and visual features.
We train FoCo on the Conceptual Captions 3M (CC3M)~\cite{sharma2018conceptual} with short captions from~\cite{DreamLIP}. 
For each caption, we further employ the Llama-3.1-8B-Instruct model~\cite{grattafiori2024llama} to generate three to five pairs of local captions and contextual captions, which provide fine-grained supervision for the two proxy tasks.
More details about the training data and additional results with other backbones (e.g., BLIP~\cite{li2022blip}) are provided in the Appendix Sec. A.
Training proceeds in two stages: we first pretrain $F_{\text{Aggr}}$ for 3 epochs with a learning rate of $5e^{-4}$, and then jointly optimize $F_{\text{Aggr}}$ and $F_{\text{Comp}}$ for 30 epochs under the unified contrastive objective, using a batch size of 512 and learning rates of $5e^{-5}$ and $3e^{-5}$, respectively. With ViT-L/14, FoCo trains 38M parameters in 17h and infers in 0.034s/query, using LLM decomposition only as cached preprocessing.

\noindent{\textbf{Evaluation Benchmarks and Metrics.}}
We evaluate our method on four ZS-CIR benchmarks. (1) FashionIQ~\cite{wu2021fashion} contains three fashion categories (Dress, Shirt, Toptee), each with human-annotated relative captions; we report Recall@10 and Recall@50 following the Original-Split protocol. 
(2) CIRR~\cite{liu2021image} features real-world image pairs with natural language modifications and visually similar distractors; we report Recall@1, 5, 10, and 50 on the official test set. 
(3) CIRCO~\cite{baldrati2023zero} is a large-scale benchmark built on the COCO dataset, providing multiple ground-truths per query with a 120K-image gallery; we report mAP@5, 10, 25, and 50. 
(4) GeneCIS~\cite{vaze2023genecis} is a benchmark designed to evaluate conditional image similarity in a zero-shot manner, featuring four tasks that require models to focus on or change specific attributes or objects in an image based on textual conditions. We report Recall@1, 2, and 3 for all four task types.

\noindent{\textbf{Comparison with State-of-the-art Methods.}}
We compare FoCo against a suite of representative ZS-CIR baselines. Pic2Word~\cite{saito2023pic2word} maps reference images into pseudo-word tokens in the CLIP token space. SEARLE-XL~\cite{baldrati2023zero} integrates such tokens into GPT-generated captions. LinCIR~\cite{gu2024language} masks subject entities in captions for efficiency-aware training. 
Context-I2W~\cite{tang2024context} focuses on extracting visual cues aligned with the text, while PrediCIR~\cite{tang2025missing} enhances pseudo-tokens by inferring missing visual semantics from the caption.
MoA~\cite{li2025rethinking} introduces multi-object pseudo-token learning, while HIT~\cite{li2025hierarchy} learns multiple pseudo-words organized across hierarchical semantic levels.
SlerpTAT~\cite{jang2024spherical} leverages LoRA-based tuning of the image encoder for additive composition. 
For fair comparison, we report results on ViT-L/14 for all baselines and further evaluate generalization of FoCo on ViT-G/14.
We further compare with recent training-free and LLM-based ZS-CIR methods, including CoTMR, IP-CIR, and PDV, in the Appendix.
\begin{table*}[t!]
\newcolumntype{C}{>{\centering\arraybackslash}S[table-format=2.1]}
    \sisetup{detect-weight=true, detect-inline-weight=math, input-symbols={-}}
  \centering
  \caption{Retrieval results on CIRR and CIRCO test sets. Best scores are in bold.}
  \renewcommand{\arraystretch}{0.9}
  \setlength{\tabcolsep}{1.5pt}
  \begin{adjustbox}{width=0.85\textwidth}
  \begin{tabular}{
    >{\centering\arraybackslash}p{1.8cm}
    l
    *{4}{C}
    @{\hspace{6pt}\sepline\hspace{6pt}}
    *{4}{C}
  }
    \toprule
    \multirow{4}{*}{\centering Backbones} & \multirow{4}{*}{\centering Methods} 
    & \multicolumn{4}{c@{\hspace{12pt}}}{CIRR} 
    & \multicolumn{4}{c}{CIRCO} \\
    \cmidrule(lr){3-6} \cmidrule(lr){7-10}
    & & \multicolumn{4}{c@{\hspace{12pt}}}{Recall@} 
      & \multicolumn{4}{c}{mAP@} \\
    & & \multicolumn{1}{c}{1} 
      & \multicolumn{1}{c}{5} 
      & \multicolumn{1}{c}{10} 
      & \multicolumn{1}{c@{\hspace{12pt}}}{50} 
      & \multicolumn{1}{c}{5} 
      & \multicolumn{1}{c}{10} 
      & \multicolumn{1}{c}{25} 
      & \multicolumn{1}{c}{50} \\
    \midrule
    \multirow{9}{*}{\centering ViT-L/14}
    & Pic2Word~\cite{saito2023pic2word} \conf{CVPR'23} & 23.9 & 51.7 & 65.3 & 87.8 & 8.7 & 9.5 & 10.6 & 11.3 \\
    & SEARLE-XL~\cite{baldrati2023zero} \conf{ICCV'23} & 24.2 & 52.5 & 66.3 & 88.8 & 11.7 & 12.7 & 14.3 & 15.1 \\
    & Context-I2W~\cite{tang2024context} \conf{AAAI'24} & 25.6 & 55.1 & 68.5 & 89.8 & 13.0 & 14.6 & 16.1 & 17.2 \\
    & LinCIR~\cite{gu2024language} \conf{CVPR'24} & 25.0 & 53.3 & 66.7 & {--} & 12.6 & 13.6 & 15.0 & 15.9 \\
    & SlerpTAT~\cite{jang2024spherical} \conf{ECCV'24} & 30.9 & 59.4 & 70.9 & 89.2 & 17.0 & 17.8 & 19.6 & 20.6 \\
    & PrediCIR~\cite{tang2025missing} \conf{CVPR'25} & 27.2 & 57.0 & 70.2 & {--} & 15.7 & 17.1 & 18.6 & 19.3 \\
    & MoA~\cite{li2025rethinking} \conf{SIGIR'25} & 27.1 & 56.5 & 69.2 & 90.0 & 15.3 & 17.1 & 18.5 & 19.3 \\
    & HIT~\cite{li2025hierarchy} \conf{ICCV'25} & 27.9 & 57.6 & 70.5 & 90.4 & 15.5 & 16.7 & 18.9 & 19.9 \\
    & \textbf{FoCo} & \textbf{34.5} & \textbf{65.4} & \textbf{74.7} & \textbf{92.1} & \textbf{17.5} & \textbf{18.1} & \textbf{19.9}& \textbf{20.8} \\
    \midrule
    \multirow{3}{*}{\centering ViT-G/14}
    & LinCIR~\cite{gu2024language} \conf{CVPR'24} & 35.3 & 64.7 & 76.1 & {--} & 19.7 & 21.0 & 23.1 & 24.2 \\ 
    & PrediCIR~\cite{tang2025missing} \conf{CVPR'25} & 37.0 & 66.1 & 77.9 & {--} & 23.7 & 24.6 & 25.4 & 26.0 \\
    & \textbf{FoCo} & \textbf{38.5} & \textbf{68.6} & \textbf{79.7} & \textbf{95.0} & \textbf{25.5} & \textbf{27.0} & \textbf{27.8} & \textbf{28.9} \\
    \bottomrule
  \end{tabular}
  \end{adjustbox}
  \label{tab:cirr-circo}
\end{table*}

\subsection{Main Results}
\noindent\textbf{FashionIQ.} 
On this benchmark, which focuses on domain-specific attribute edits such as color or material changes, 
FoCo achieves superior results across most metrics (\cref{tab:fiq-only}).
This indicates its ability to integrate fine-grained visual attributes into coherent queries. Unlike CIRR, which emphasizes object-level changes, FashionIQ stresses subtle property modifications, and FoCo demonstrates strong generalization across different types of changes.

\begin{table*}[t!]
  \centering
  \small
  \caption{Quantitative results on the GeneCIS benchmark across four task types: Focus, Change, Attribute, and Object. More results are in App. A.}
  \label{tab:genecis}
  \sisetup{detect-weight=true, detect-inline-weight=math, input-symbols={-}}
  \renewcommand{\arraystretch}{1.0}
  \setlength{\tabcolsep}{2pt}
  \begin{adjustbox}{width=0.95\textwidth}
  \begin{tabular}{
    >{\centering\arraybackslash}p{2cm}
    l
    *{12}{C}
  }
    \toprule
    \multirow{3}{*}{Backbones} & \multirow{3}{*}{Methods} & \multicolumn{3}{c}{Focus} & \multicolumn{3}{c}{Change} & \multicolumn{3}{c}{Attribute} & \multicolumn{3}{c}{Object} \\
    \cmidrule(lr){3-5} \cmidrule(lr){6-8} \cmidrule(lr){9-11} \cmidrule(lr){12-14}
    & & {R1} & {R2} & {R3} & {R1} & {R2} & {R3} & {R1} & {R2} & {R3} & {R1} & {R2} & {R3} \\
    \midrule
    \multirow{6}{*}{\centering \makecell{CLIP-\\ ViT-L/14}}
    & Pic2Word~\cite{saito2023pic2word} \conf{CVPR'23} & 12.1 & 23.1 & 32.3 & 10.3 & 19.9 & 28.6 & 14.8 & 26.5 & 35.9 & 7.6 & 16.6 & 24.9 \\
    & SEARLE-XL~\cite{baldrati2023zero} \conf{ICCV'23} & 12.4 & 23.2 & 33.0 & 12.2 & 21.1 & 29.6 & 16.7 & 27.5 & 37.4 & 7.9 & 16.8 & 25.2 \\
    & Context-I2W~\cite{tang2024context} \conf{AAAI'24} & 13.0 & 24.2 & 34.3 & 12.1 & 22.2 & 31.3 & 16.8 & 29.4 & 39.4 & 8.2 & 17.0 & 26.2 \\
    & LinCIR~\cite{gu2024language} \conf{CVPR'24} & 12.6 & 23.7 & 33.9 & 11.8 & 21.9 & 30.9 & 16.6 & 29.0 & 39.2 & 7.9 & 16.6 & 25.6 \\
    & PrediCIR~\cite{tang2025missing} \conf{CVPR'25} & 15.5 & 25.5 & 36.9 & 17.8 & 28.0 & 34.8 & 18.5 & 31.2 & 39.0 & 14.8 & 22.3 & 32.7 \\
    & \textbf{FoCo (ours)} & \textbf{17.9} & \textbf{26.9} & \textbf{37.2} & \textbf{17.9} & \textbf{28.1} & \textbf{35.7} & \textbf{19.3} & \textbf{31.5} & \textbf{39.7} & \textbf{16.6} & \textbf{23.5} & \textbf{33.2} \\
    \midrule
    \multirow{3}{*}{\centering \makecell{CLIP-\\ ViT-G/14}}
    & LinCIR~\cite{gu2024language} \conf{CVPR'24} & 14.6 & 26.1 & 35.2 & 12.8 & 23.3 & 31.9 & 18.4 & 31.6 & 40.2 & 9.0 & 17.7 & 26.9 \\
    & PrediCIR~\cite{tang2025missing} \conf{CVPR'25} & 16.1 & 26.3 & 37.5 & \textbf{19.4} & 31.5 & 39.8 & 19.6 & 32.0 & 40.8 & 15.9 & 25.8 & 36.5 \\
    & \textbf{FoCo (ours)} & \textbf{18.4} & \textbf{27.9} & \textbf{39.0} & 19.3 & \textbf{31.8} & \textbf{40.0} & \textbf{19.6} & \textbf{32.3} & \textbf{41.0} & \textbf{18.1} & \textbf{27.4} & \textbf{38.0} \\
    \bottomrule
  \end{tabular}
\end{adjustbox}
\end{table*}

\noindent\textbf{CIRR.} FoCo surpasses all existing ZS-CIR baselines on CIRR, improving R@1 by 3.6 points over SlerpTAT on ViT-L/14 (\cref{tab:cirr-circo}). CIRR contains natural images with complex scenes and multiple objects, making localized edits especially challenging. Models that rely on global feature fusion or naive pseudo-token composition often fail to distinguish visually similar distractors. While methods like PrediCIR and HIT~\cite{li2025hierarchy} attempt to mitigate noise in pseudo-token mapping, our results show that explicit localization leads to more accurate disambiguation.

\noindent\textbf{CIRCO.} With a 120K-image gallery, CIRCO introduces significant retrieval noise. FoCo consistently leads in mAP metrics, reflecting the quality of its composed features (\cref{tab:cirr-circo}). On ViT-L/14, it achieves 17.5 mAP@5, outperforming SlerpTAT and PrediCIR by 0.5 and 1.8 points, respectively. These results validate the effectiveness of the proposed context-guided completion module in generating robust and precise query representations. 

\noindent\textbf{GeneCIS.} This benchmark evaluates conditional similarity across four task types that combine focus/change intents with attribute/object granularities. FoCo achieves the highest scores (\cref{tab:genecis}), outperforming PrediCIR most notably on Focus (+2.4 Recall@1) and Object (+1.8 Recall@1) tasks, where precise grounding of the condition-specified visual region is most critical.
In contrast to pseudo-token concatenation, FoCo explicitly models both what to modify and how to compose, resulting in a more interpretable composition process.
Collectively, these results demonstrate that a learnable, two-stage formulation yields superior generalization across diverse modification types compared to static pseudo-token baselines, without requiring triplet supervision.

\subsection{Ablation Study and Analysis}

\noindent{\textbf{Proxy Task.}}
FoCo learns an adaptive composition mechanism within the CLIP semantic space. Specifically, $F_{\text{Aggr}}$ grounds modification-relevant visual evidence, while $F_{\text{Comp}}$ performs a dynamic, text-driven semantic transformation. Removing either module leads to an ill-posed proxy task.
To evaluate them independently, we follow~\cite{saito2023pic2word} and replace $F_{\text{Comp}}$ with linear feature fusion (1b), and use the class-activation-based patch masking strategy from~\cite{zhang2024zero} to provide coarse visual evidence to $F_{\text{Comp}}$ (1c).
As shown in~\cref{tab:ablation}, the naive global baseline (1a) performs poorly. Using only $F_{\text{Aggr}}$ (1b) significantly improves accuracy via meaningful grounding, while using only $F_{\text{Comp}}$ (1c) yields smaller gains due to coarse localized activations. This gap is particularly pronounced in CIRR, which requires precise object-level localization, compared to the attribute-centric FashionIQ.
Overall, two proxy tasks decouple the composition process: aggregation determines what to modify, while completion defines how to modify.

\begin{table}[t!]
    \centering
    \renewcommand{\arraystretch}{0.9}
    \setlength{\tabcolsep}{2pt}
    \begin{minipage}[c]{0.48\linewidth}
        \centering
        \begin{adjustbox}{width=\linewidth}
        \begin{tabular}{l ccc cc}
        \toprule
        \multirow{2}{*}{\textbf{Experiment}} & \multicolumn{3}{c}{\textbf{CIRR}} & \multicolumn{2}{c}{\textbf{FashionIQ}} \\
        \cmidrule(lr){2-4} \cmidrule(lr){5-6}
        & R1 & R5 & R10 & R10 & R50 \\
        \toprule
        \textbf{full model} & \textbf{34.5} & \textbf{65.4} & \textbf{74.7} & \textbf{31.5} & \textbf{53.2} \\
        \midrule
        1a.\;\;image+text   & 12.4  & 36.2 & 49.1 & 19.8 & 35.7\\
        1b.\;\;aggregation-only & 25.8 & 54.6 & 67.1 & 26.4 & 46.9 \\
        1c.\;\;completion-only & 22.2 & 49.8 & 60.9 & 23.5 & 42.7 \\
        \midrule
        2a.\;\;InfoNCE loss & 33.1 & 63.9 & 73.2 & 30.2 & 51.7 \\
        2b.\;\;naive neg. (aggr) & 31.0 & 61.8 & 71.5 & 28.8 & 50.4 \\
        2c.\;\;naive neg. (comp) & 32.3 & 62.6 & 72.4 & 27.9 & 49.1 \\
        2d.\;\;w/o joint $\mathcal{L}^{\text{aggr}}$ & 33.2 & 63.5 & 73.0 & 30.4 & 52.0 \\
        \bottomrule
        \end{tabular}
        \end{adjustbox}
    \end{minipage}
    \hfill
    \begin{minipage}[c]{0.48\linewidth}
        \centering
        \begin{adjustbox}{width=\linewidth}
        \begin{tabular}{l ccc cc}
        \toprule
        \multirow{2}{*}{\textbf{Experiment}} & \multicolumn{3}{c}{\textbf{CIRR}} & \multicolumn{2}{c}{\textbf{FashionIQ}} \\
        \cmidrule(lr){2-4} \cmidrule(lr){5-6}
        & R1 & R5 & R10 & R10 & R50 \\
        \toprule
        \textbf{full model} & \textbf{34.5} & \textbf{65.4} & \textbf{74.7} & \textbf{31.5} & \textbf{53.2} \\
        \midrule
        3a.\;\;rule-based decom. & 30.7 & 60.9 & 70.6 & 28.5 & 49.6 \\
        3b.\;\;single local caption & 32.6 & 63.1 & 72.7 & 30.0 & 51.1 \\
        \midrule
        4a.\;\;w/o $\odot$ query & 33.5 & 64.1 & 73.4 & 30.7 & 52.1 \\
        4b.\;\;dot-prod attn & 33.0 & 63.7 & 73.0 & 30.3 & 51.6 \\
        \midrule
        5a.\;\;w/o $\mathcal{T(\cdot)}$ & 32.8 & 63.0 & 72.5 & 27.5 & 47.9 \\
        5b.\;\;w/o $\bar{v}_{i,i_k}$ & 33.1 & 63.5 & 73.0 & 30.2 & 51.5 \\
        5c.\;\;w/o query init & 32.9 & 63.2 & 72.8 & 29.9 & 51.0 \\
        \bottomrule
        \end{tabular}
        \end{adjustbox}
    \end{minipage}
    \caption{Ablation study on CIRR and FashionIQ. The table is split into two columns for better visualization. Both sides compare against the full model.}
    \label{tab:ablation}
\end{table}

\noindent{\textbf{Component Analysis.}}
We first ablate our training strategies and architectural choices in~\cref{tab:ablation}. 
For the \textit{contrastive objective}, replacing our multi-positive loss with InfoNCE (\textbf{2a}) drops performance. Using naïve intra-image negatives (\textbf{2b, 2c}) weakens cross-instance discrimination, and removing the aggregation loss during joint training (\textbf{2d}) confirms $F_{\text{Aggr}}$ benefits from continued explicit supervision.
Regarding \textit{architectural designs}, removing the text-guided query (\textbf{4a}) weakens localization, as it lacks global visual context and cannot disambiguate expressions whose referents differ across images. Replacing the cross-attention layer with a dot-product attention (\textbf{4b}) also reduces performance, highlighting the need for deeper cross-modal interaction. 
For the completion module, removing the Transformer blocks (\textbf{5a}) disrupts cross-modal interaction, leading to a performance drop. 
Removing the $\bar{v}_{i,i_k}$ from the key/value sequence (\textbf{5b}) also degrades results, confirming that explicitly supplying unattended visual context helps the model preserve semantics that should remain unchanged.
Replacing the visual-anchor initialization with a random learnable token (\textbf{5c}) also degrades performance, confirming that the localized visual subject $v^{\text{loc}}_{i,i_k}$ provides a necessary semantic anchor for extracting modification-relevant information.

\begin{figure}[t!]
    \centering
    \begin{minipage}[c]{0.38\linewidth} 
        \centering
        \includegraphics[width=\linewidth]{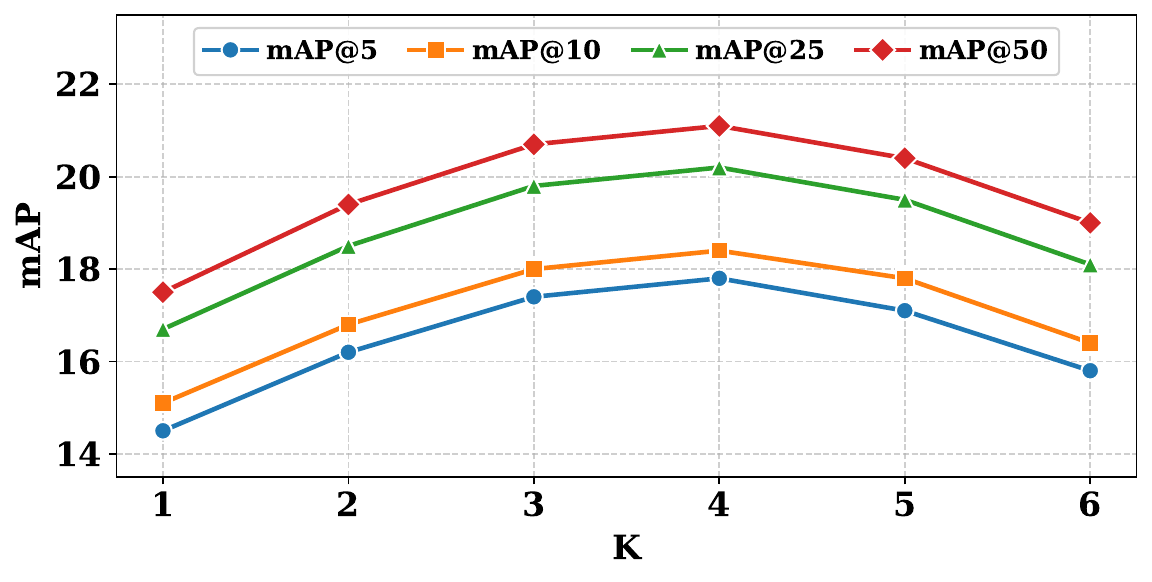}
        \includegraphics[width=\linewidth]{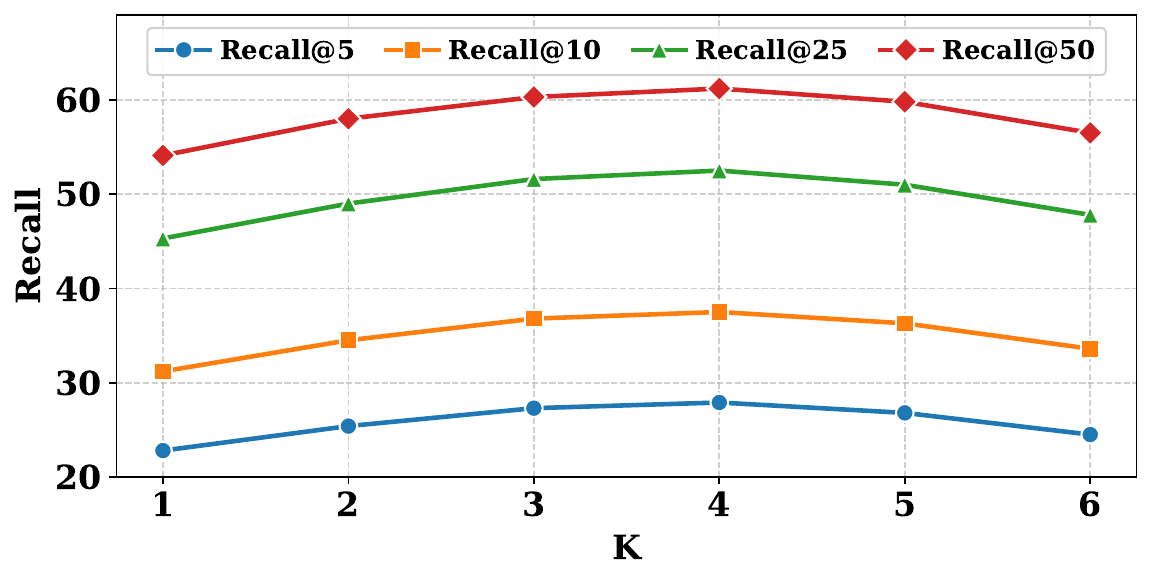}
        \caption{Impact of local captions number ($K$) on CIRCO test set.}
        \label{fig:k_sensitivity}
    \end{minipage}
    \hfill
    \begin{minipage}[c]{0.6\linewidth} 
        \centering
        \includegraphics[width=\linewidth]{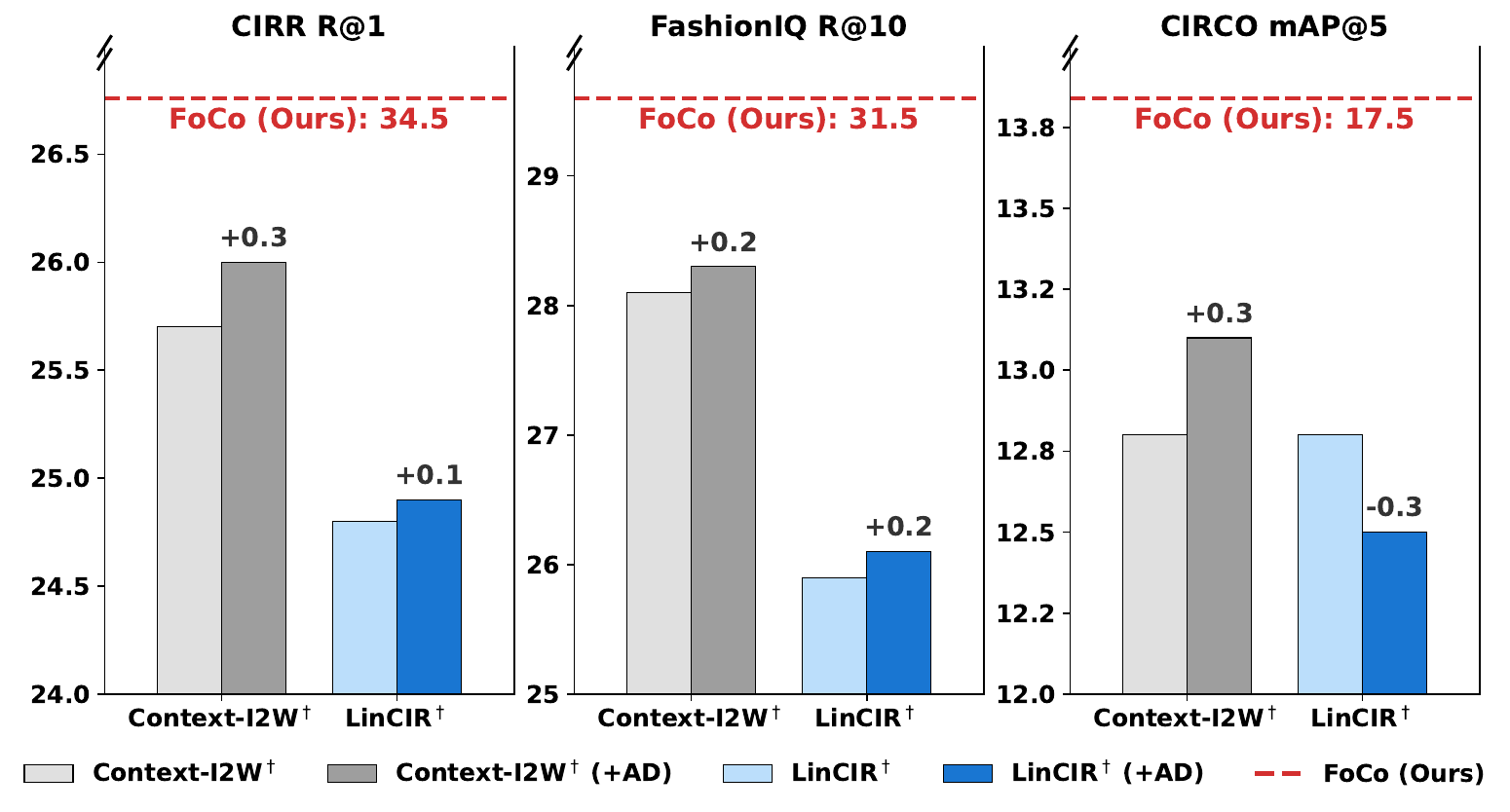}
        \caption{Effect of augmenting baseline training sets with our LLM-reconstructed captions (+AD). $^\dagger$ indicates our reproduction using the official implementation. Full results are in Appendix.}
        \label{fig:data_ablation}
    \end{minipage}
\end{figure}

\noindent{\textbf{Analysis on Textual Decomposition.}}
The quality and granularity of our text decomposition are crucial. 
Replacing LLM-decomposed descriptions with rule-based syntactic extraction (\textbf{3a}) causes significant degradation, as rigid grammatical splitting creates isolated fragments and ungrammatical contexts, disrupting semantic coherence.
Furthermore, relying on a single local caption per image (\textbf{3b}, i.e., $K\!=\!1$) noticeably reduces accuracy. This highlights that multi-view training pairs are essential to capture diverse compositional patterns. 
To investigate this systematically, we visualize the sensitivity to decomposition granularity ($K$) on the large-scale CIRCO dataset in~\cref{fig:k_sensitivity}. 
Consistent with \textbf{3b}, insufficient decomposition leads to the lowest performance.
As $K$ increases, performance steadily improves and peaks at $K\!=\!4$.  
Over-decomposition (\textbf{$K\!\ge\!6$}) causes performance to decline, as it compels the LLM to generate hallucinated visual details or excessive semantic noise. These results confirm that a dynamic \textbf{$K\!\in\![3,5]$} provides the optimal balance between fine-grained supervision and semantic fidelity.

\noindent{\textbf{Impact of Data Decomposition.}}
A natural concern is whether the LLM decomposition introduces extra semantic information that drives FoCo's gains.
To test this, we concatenate the decomposed local captions and contextual captions back into complete sentences and append them as additional training pairs to two baselines (denoted ``+AD'').
As shown in~\cref{fig:data_ablation}, the additional data brings only marginal or even negative effects: Context-I2W$^\dagger$ (+AD) improves by at most +0.3, while LinCIR$^\dagger$ (+AD) actually \emph{drops} on CIRCO mAP@5 ($-0.3$), likely because its text-only training regime is sensitive to the distributional shift introduced by the restructured captions.
In contrast, FoCo surpasses all augmented baselines by a large margin, confirming that our gains originate from the proxy-task design rather than extra training data (full results in Appendix).

\subsection{Visualization and Analysis}
\label{sec:visualization}
\begin{figure}[t!]
    \centering 
    \includegraphics[width=0.9\linewidth]{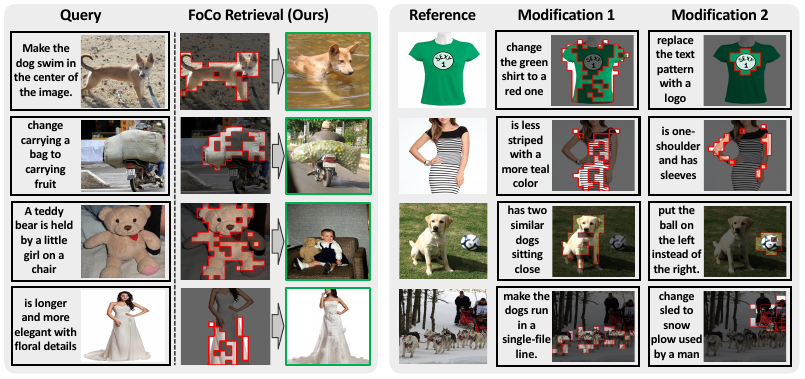}
    \caption{Qualitative visualization of FoCo. Red regions indicate text-anchored high-response areas. Left: FoCo retrieves images consistent with the modification text. Right: Different modification texts on the same reference yield different focus responses.}
    \label{fig:vis}
\end{figure}

To understand how FoCo performs modification-guided image retrieval, we visualize the activation maps from the text-anchored visual aggregation module $F_{\text{Aggr}}$ alongside the retrieval results in \cref{fig:vis}. 
Each query includes a reference image and modification text, and the red-highlighted patches indicate the regions most relevant to the modification. 
On the right, multiple modification texts are applied to the same reference image. This reveals how FoCo dynamically adjusts its focus as the intended change varies. Attribute-level edits (e.g., changing color or texture) result in fine-grained and localized activation patterns, while object-level or relation-level modifications lead to broader spatial activation that covers the relevant entity or configuration. These consistent and meaningful shifts indicate that FoCo does not rely on a fixed composition rule, but actively restructures its visual grounding conditioned on the modification text. $F_{\text{Comp}}$ then integrates these focused representations with global context to form a coherent final representation. In summary, FoCo localizes the modification-relevant regions and integrates them into a target representation in a controllable manner.
More retrieval examples are shown in Figs. 1 and 2 in the Appendix.

\noindent{\textbf{Analysis.}} 
A key challenge in ZS-CIR is the inherent domain gap between descriptive training captions and instruction-driven modification texts.
We argue that FoCo mitigates this gap through its adaptive composition mechanism, which modulates visual features within the CLIP semantic space according to textual conditions. Leveraging the pretrained CLIP text encoder, FoCo also inherits linguistic robustness across varied input forms.
Specifically, $F_{\text{Aggr}}$ first grounds modification-relevant visual evidence, whether the intent is to add, modify, or remove. 
$F_{\text{Comp}}$ then integrates these features with contextual cues to complete the transformation.
This is supported by our decomposition strategy, which provides non-overlapping localized and contextual features, while the multi-view training pairs expose the model to varying visual-semantic dependencies ranging from small object preservation to large scene shifts.
Moreover, the cross-instance contrastive objective prevents textual shortcuts, forcing the model to discern the necessary visual contribution based on these dependencies. 
Consequently, as shown in~\cref{fig:vis}, this design enables FoCo to dynamically modulate evidence in response to diverse instructions.

\section{Conclusion}
We introduce Focus-then-Complete (FoCo), a framework for zero-shot composed image retrieval that treats visual–textual composition as a learnable procedure. FoCo first isolates visual cues relevant to the textual modification and then completes the target semantics through contextual reasoning. These two stages correspond to two proxy tasks, text-anchored visual aggregation and context-conditioned semantic completion, trained jointly with a cross-instance contrastive objective. This design yields accurate, stable, and interpretable compositional representations without relying on triplet supervision. 
We believe FoCo offers a clear and effective foundation for learning controllable multimodal composition~\cite{hua2024mmcomposition}, and we plan to extend it to more complex, open-ended vision–language settings~\cite{ossowski2024prompting, li2025unveiling}.

\section*{Acknowledgments}
This work was supported by the National Natural Science Foundation of China under Grants 62576329 and 62336001.

\bibliographystyle{splncs04}
\bibliography{main_foco}
\end{document}